\documentclass[conference]{IEEEtran}
\usepackage{cite}
\usepackage{amsmath,amssymb,amsfonts}
\usepackage{graphicx}
\usepackage{textcomp}
\usepackage[table]{xcolor}
\usepackage{booktabs,array,multirow}
\usepackage{url}
\usepackage{microtype}
\usepackage{makecell}

\newsavebox{\PNmaintablebox}
\newcommand{\PNtableonefont}{\scriptsize}
\makeatletter
\newcommand{\PNrecordtablefont}{%
  \Gscale@div\PNtablescale{\columnwidth}{\wd\PNmaintablebox}%
  \dimen@=\f@size pt\relax
  \dimen@=\PNtablescale\dimen@
  \@tempdima=\PNtablescale\baselineskip
  \xdef\PNtableonefont{\noexpand\fontsize
    {\the\dimen@}{\the\@tempdima}\noexpand\selectfont}%
}
\makeatother

\definecolor{PNpurple}{HTML}{B5ADFD}
\definecolor{PNlight}{HTML}{E7E5FF}
\definecolor{PNgain}{HTML}{6F139A}
\definecolor{PNgroup}{HTML}{F2F2F2}
\newcommand{\gain}[1]{{\color{PNgain}\scriptsize\ensuremath{\uparrow #1}}}

\newcommand{\ind}{\mathbb{I}}

\newcommand{\pos}{\mathrm{positive}}
\newcommand{\negat}{\mathrm{negative}}
\newcommand{\lead}[1]{\noindent\textbf{#1.}\ }
\begin{document}

\title{Tiny Evidence Region Localization with Selective Contrastive On-Policy Self-Distillation}

\author{Yujuan Pang, Wenhao Xu, Yinglei Teng\\
\textit{Beijing University of Posts and Telecommunications}\\
{Beijing, China}}
\maketitle

\begin{abstract}
Multimodal large language models have made progress in general visual tasks. However, limitations remain in fine-grained visual tasks requiring evidence from minute image regions. Existing methods primarily train models using evidence-related regions while overlooking the influence of extensive background regions, resulting in a limited ability to resist interference from irrelevant regions. We propose selective positive-negative on-policy self-distillation (SPN-OPSD), which considers both evidence and background regions. Supervision from both views improves the model’s ability to resist interference from irrelevant regions. Furthermore, selective token-level distillation provides more informative supervision, improving the model’s ability to attend to evidence regions in Fine-Grained Visual Tasks. On Qwen3.5-2B, SPN-OPSD improves accuracy on V* Bench, ZoomBench, and MME-RealWorld-Lite from 76.44\%, 45.92\%, and 57.43\% to 84.82\%, 54.56\%, and 59.56\%, respectively. Similar gains are also observed on Qwen3.5-0.8B.
\end{abstract}

\begin{IEEEkeywords}
fine-grained visual tasks, multimodal large language models, on-policy self-distillation
\end{IEEEkeywords}

\section{Introduction}
\label{sec:intro}
Multimodal large language models (MLLMs) have demonstrated strong capabilities in general visual understanding and reasoning~\cite{r01,r02,r03}. However, they still face challenges in fine-grained visual tasks, where answers often depend on small but crucial details that occupy only a small portion of an image~\cite{r04,r05}. Extensive backgrounds and similar objects further complicate the identification of relevant evidence.

The recent “Thinking with Images” paradigm acquires local evidence by iteratively zooming into and cropping regions of interest during inference~\cite{r06,r07,r08}. Other methods develop the ability to perceive local visual details through local-view supervision during training~\cite{r09,r10}, or enhance fine-grained visual perception through visual evidence localization~\cite{r11,r12} and attention alignment~\cite{r13}. These methods indeed improve model capabilities by teaching models ``what to attend to.'' However, inspired by the observation that background regions can also introduce interference~\cite{r06,r14}, we argue that it is also necessary to teach models ``what not to attend to.''

To enhance fine-grained visual perception, DeFacto~\cite{r11} constructs counterfactual examples by masking evidence regions and trains models to output “Unknown” when crucial evidence is absent. VA-OPD~\cite{r15} defines Visual Advantage as the difference in teacher token log-probabilities with and without fine-grained visual details, and uses it to adjust distillation weights. Despite their progress on fine-grained visual tasks, these methods have a limitation: they use counterfactual signals to learn abstention behavior or adjust positive distillation weights, but do not explicitly guide the student away from predictive distributions formed in the absence of evidence. They therefore lack a direct constraint on erroneous predictions supported by nonessential visual information.

\begin{figure}[t]
    \centering
    \includegraphics[width=\columnwidth]{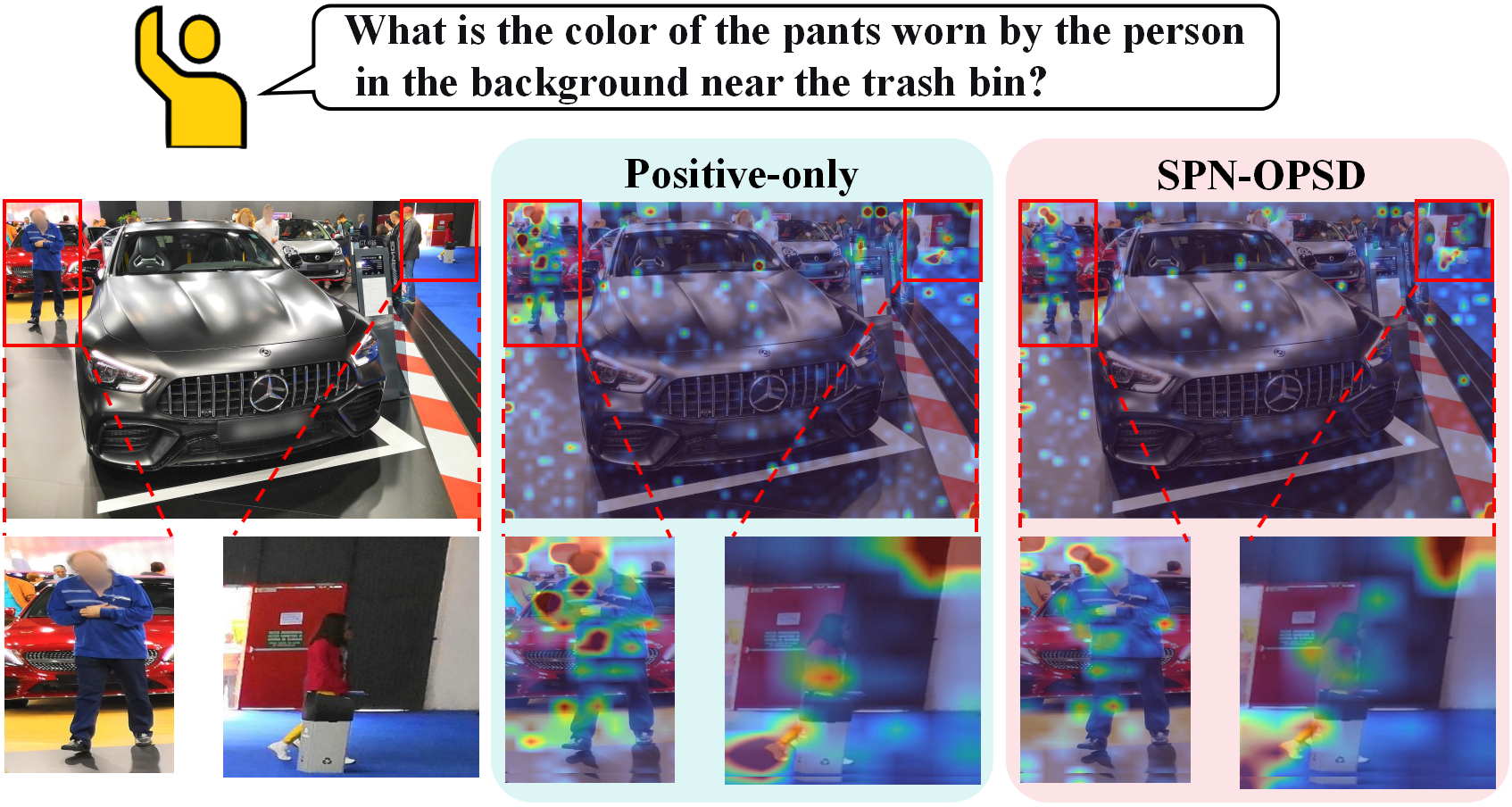}
    \caption{Visual Attention Visualization}
    \label{fig:attention}
\end{figure}

To address this limitation, we propose selective positive-negative on-policy self-distillation (SPN-OPSD), a positive–negative teacher framework for on-policy self-distillation that uses both evidence and background regions as privileged information. Specifically, the positive teacher takes a cropped view of the evidence region as input and provides a target distribution grounded in local evidence. The negative teacher takes a view in which crucial evidence is suppressed while background information is retained, capturing predictive tendencies in the absence of crucial evidence. The student takes the original image as input and receives selective negative supervision while continuously learning from the positive teacher’s distribution. This mechanism enables the student to learn to use crucial visual evidence while selectively moving away from predictive distributions unsupported by that evidence. As shown in Fig.~\ref{fig:attention}, the student trained with SPN-OPSD maintains its focus on the evidence region and devotes less attention to irrelevant regions than its counterpart trained with positive-only distillation, demonstrating a greater ability to resist interference from irrelevant image regions. 

In summary, our contributions are as follows:
\begin{itemize}

\item We introduce SPN-OPSD for fine-grained visual tasks. Complementary evidence-enhanced and evidence-suppressed views incorporate both relevant evidence and background interference into the distillation process.
\item We develop a token-level selection mechanism based on the distributions of the student and both teachers. Teacher disagreement identifies valuable supervision positions, while the student's relative difference to each teacher identifies predictions susceptible to background interference. The student approaches the positive teacher at all valid positions and moves away from the negative teacher at selected positions, promoting evidence use while reducing irrelevant interference.
\item We validate the effectiveness of our method on three benchmarks and at two model sizes. Our method outperforms both existing positive-evidence methods and the evaluated open-source models.
\end{itemize}

\begingroup
\scriptsize
\setlength{\tabcolsep}{3pt}
\renewcommand{\arraystretch}{1.08}
\sbox{\PNmaintablebox}{%
\begin{tabular}{@{}lccccc@{}}
\toprule
\multicolumn{1}{@{}c}{Model}
& Size
& V* Bench
& ZoomBench
& \makecell[c]{MME-\\RealWorld-Lite}
& Avg \\
\midrule

\rowcolor{PNgroup}
\multicolumn{6}{c}{\textit{Visual agents}} \\
DeepEyes V2 & 7B & 81.68 & 44.97 & 51.59 & 59.41 \\
Thyme & 7B & 82.20 & 45.09 & 55.18 & 60.82 \\
\midrule

\rowcolor{PNgroup}
\multicolumn{6}{c}{\textit{Closed-source models}} \\
GPT-5.2 & -- & 79.06 & 50.89 & 47.06 & 59.00 \\
GPT-5.4 & -- & 76.96 & \underline{52.66} & 48.10 & 59.24 \\
\midrule

\rowcolor{PNgroup}
\multicolumn{6}{c}{\textit{Open-source models}} \\
Qwen3.5 & 4B & \underline{84.29} & 47.69 & 35.80 & 55.93 \\
Qwen3.5 & 9B & 82.72 & 52.07 & 46.12 & 60.30 \\
MiniCPM-V-4.5 & 9B & 70.68 & 42.60 & 42.31 & 51.86 \\
MiMo-V2.5 & 310B & 82.20 & 51.24 & 46.69 & 60.04 \\
\midrule

\rowcolor{PNgroup}
\multicolumn{6}{c}{\textit{Matched backbone and training data}} \\
Qwen3.5 & 0.8B & 66.49 & 40.71 & 50.65 & 52.62 \\
+ VA-OPD & 0.8B & 64.40 & 39.29 & 46.95 & 50.21 \\
+ DeFacto & 0.8B & 67.54 & 42.84 & 51.75 & 54.04 \\
+ CompoDistill & 0.8B & 68.59 & 41.66 & 52.79 & 54.35 \\

\rowcolor{PNpurple}
+ SPN-OPSD (ours) & 0.8B & 71.73 & 45.44 & 52.16 & 56.44 \\
\rowcolor{PNpurple}
& & \textcolor{PNgain}{↑5.24}
  & \textcolor{PNgain}{↑4.73}
  & \textcolor{PNgain}{↑1.51}
  & \textcolor{PNgain}{↑3.82} \\

Qwen3.5 & 2B & 76.44 & 45.92 & 57.43 & 59.93 \\
+ VA-OPD & 2B & 83.25 & 43.55 & 53.26 & 60.02 \\
+ DeFacto & 2B & 79.58 & 49.23 & \underline{58.99} & 62.60 \\
+ CompoDistill & 2B & 82.72 & 50.53 & 58.10 & \underline{63.78} \\

\rowcolor{PNpurple}
+ SPN-OPSD (ours) & 2B
& \textbf{84.82}
& \textbf{54.56}
& \textbf{59.56}
& \textbf{66.31} \\
\rowcolor{PNpurple}
& & \textcolor{PNgain}{↑8.38}
  & \textcolor{PNgain}{↑8.64}
  & \textcolor{PNgain}{↑2.13}
  & \textcolor{PNgain}{↑6.38} \\
\bottomrule
\end{tabular}%
}
\PNrecordtablefont
\endgroup

\section{Our Proposed Method}
\label{sec:method}

\begin{figure*}[t]
\centering
\includegraphics[width=0.83\textwidth]{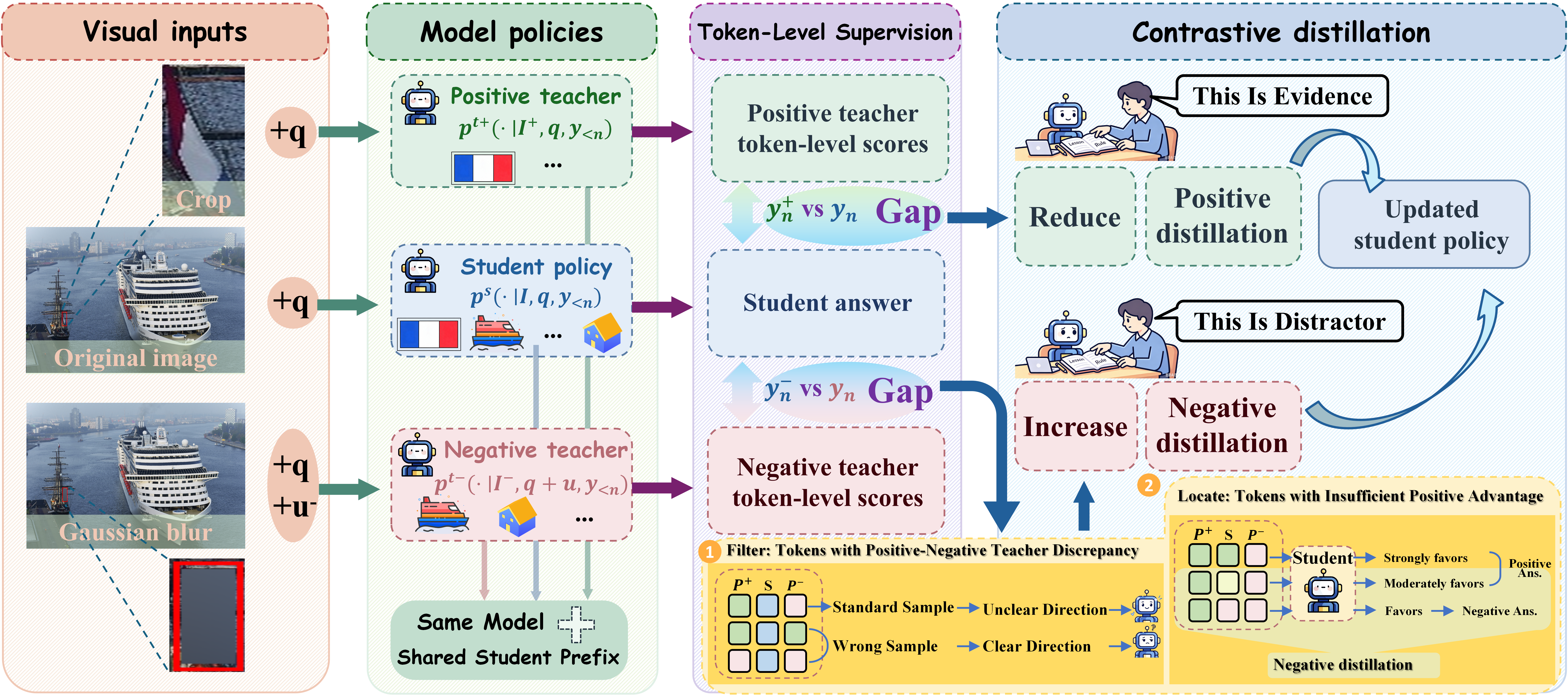}
\caption{The framework of selective positive-negative teacher distillation}
\label{fig:framework}
\end{figure*}

\subsection{Visual Views for the Student and Teachers}

Let $\mathcal D=\{(I_i,q_i,R_i)\}_{i=1}^{N}$, where $R$ is the question-relevant evidence region. We enhance and suppress the same region to obtain the positive and negative teacher views:

\begin{equation}
\begin{aligned}
I^+ &= 2\operatorname{Crop}(I,R)\\
I^-&=\operatorname{GaussianBlur}(I,R)
\end{aligned}
\label{eq:views}
\end{equation}

The crop operation extracts the evidence region, and the resulting crop is enlarged by a factor of two to emphasize local details needed to answer the question. Gaussian blur is applied only within the evidence region, weakening decisive details while preserving the surrounding background.

We instantiate the student and both teacher policies from the same MLLM $p_\theta$ under different input conditions:
\begin{equation}
\begin{aligned}
p^s(\cdot\mid I,q)&=p_\theta(\cdot\mid I,q)\\
p^{t^+}(\cdot\mid I^+,q)&=p_\theta(\cdot\mid I^+,q)\\
p^{t^-}(\cdot\mid I^-,q\oplus u^-)&=p_\theta(\cdot\mid I^-,q\oplus u^-)
\end{aligned}
\label{eq:conditions}
\end{equation}
where $\oplus$ denotes text concatenation. Because questions originate from cropped regions, we mark $R$ with a red box in the full image and append ``Answer the question by focusing only on the object inside the red bounding box'' to $q$ to avoid referential ambiguity. The negative teacher additionally receives $u^-$: ``Ignore visual information inside the red box and answer the question using only visual information outside the red box.'' The auxiliary instruction further restricts reliance on the evidence region. Enhancing and suppressing the same evidence creates complementary conditions that encourage the student to use local evidence while reducing its reliance on background information. The proposed framework is illustrated in Fig.~\ref {fig:framework}.

\subsection{Positive and Negative Distillation}
The student first generates a response
\begin{equation}
y=(y_1,\ldots,y_{|y|})\sim p^s(\cdot\mid I,q)
\label{eq:rollout}
\end{equation}

 Both teachers re-evaluate it along the same prefix $y_{<n}=(y_1,\ldots,y_{n-1})$, giving token-level distributions:

\begin{equation}
\begin{aligned}
y_n^s&=p^s(\cdot\mid I,q,y_{<n})\\
y_n^+&=p^{t^+}(\cdot\mid I^+,q,y_{<n})\\
y_n^-&=p^{t^-}(\cdot\mid I^-,q\oplus u^-,y_{<n})
\end{aligned}
\label{eq:policies}
\end{equation}
 Using the student's own generated prefixes aligns the state distributions encountered during training and inference~\cite{r16}, avoiding the prefix mismatch underlying error accumulation in off-policy distillation~\cite{r17}.

\subsubsection{Positive distillation}
The evidence crop gives the positive teacher privileged local information, so we minimize the Jensen--Shannon divergence (JSD) between the student and positive teacher at every position:
\begin{equation}
d_{s,t^+}^{(n)}=D\left(y_n^s\,\|\,y_n^+\right)
\label{eq:positive_distance}
\end{equation}
The corresponding positive distillation loss is
\begin{equation}
\mathcal{L}_{\mathrm{positive}}(\theta)
=
\mathbb{E}_{\substack{
(I,q,R)\sim\mathcal{D}\\
y\sim p^s(\cdot\mid I,q)
}}
\left[
\frac{1}{|y|}
\sum_{n=1}^{|y|}d_{s,t^+}^{(n)}
\right]
\label{eq:positive}
\end{equation}
Positive distillation covers all valid prediction positions, independently of token selection for negative supervision, and continuously preserves a clear direction for evidence learning. Compared with the sparse sequence-level rewards used in reinforcement learning with verifiable rewards (RLVR), token-level distribution matching provides denser supervision.

\subsubsection{Negative distillation}
Negative supervision is applied selectively to informative tokens through a two-stage process based on teacher disagreement and the student’s current prediction preference.

\lead{Stage 1: Selecting candidate tokens by teacher disagreement}
The teachers can produce the same prediction despite observing different views. At such positions, pulling the student toward the positive teacher while pushing it away from the negative teacher can introduce conflicting supervision. We therefore identify candidate tokens with clear teacher disagreement.

The teacher disagreement at position $n$ is
\begin{equation}
\Delta^{(n)}=D\!\left(p_n^+\,\|\,p_n^-\right)
\label{eq:disagreement}
\end{equation}
Let $v_n^\pm=\arg\max_{v\in\mathcal V}p_n^\pm(v)$ denote the teachers' top-1 predictions over vocabulary $\mathcal V$. The first-stage selection mask is
\begin{equation}
g^{(n)}=\ind[v_n^+\ne v_n^-]\,\ind[\Delta^{(n)}>\delta]
\label{eq:gate}
\end{equation}
where $\ind[\cdot]$ is an indicator function and $\delta$ is the disagreement threshold. The two conditions jointly assess prediction preference and disagreement strength. Different top-1 predictions ensure distinct preferred tokens, while the $\Delta^{(n)}$ threshold controls the magnitude of distributional disagreement.

\begin{figure}[t]
\centering
\includegraphics[width=\columnwidth]{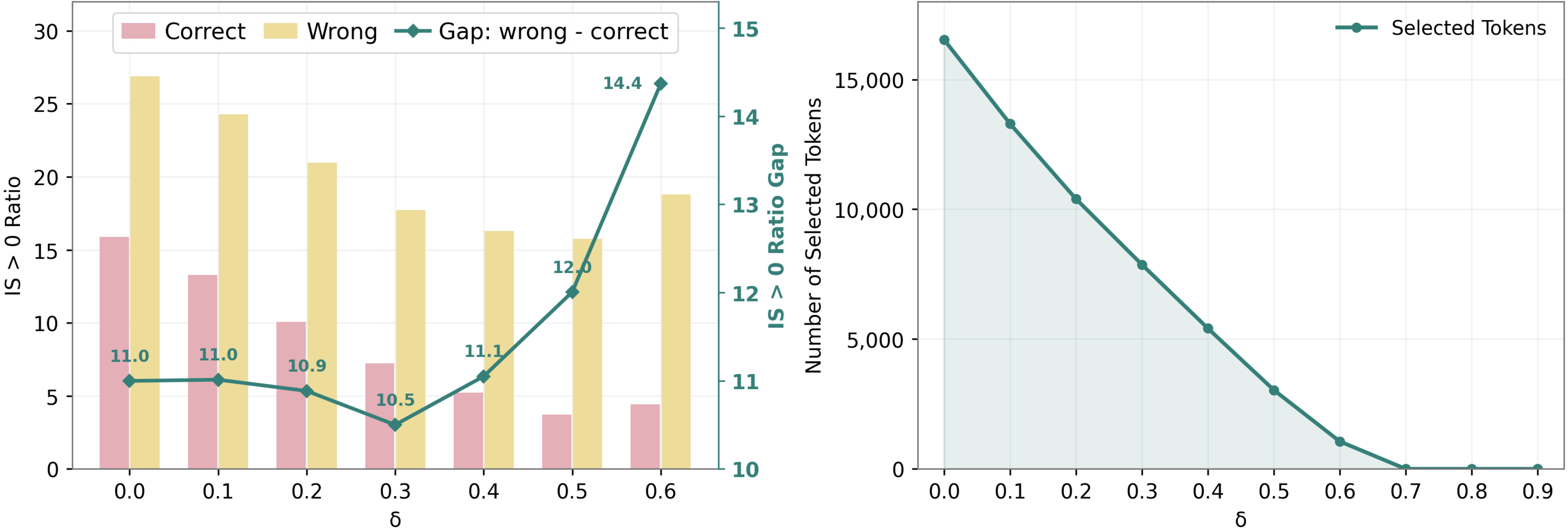}
\caption{Statistics under different teacher-disagreement thresholds}
\label{fig:delta}
\end{figure}

\noindent\textbf{How Does the Degree of Disagreement Between Positive and Negative Teachers Affect Negative Supervision?}\space
\label{sec:delta}
We examine Qwen3.5-0.8B responses on the 5.8K training examples, with both teachers evaluating the student-generated prefixes under their respective views. To characterize the student's tendency at each position, we define the interference score (IS) as
\begin{equation}
\mathrm{IS}^{(n)}=D(p_n^s\,\|\,p_n^+)-D(p_n^s\,\|\,p_n^-)
\label{eq:is}
\end{equation}
A positive IS indicates that the student is closer to the negative teacher. Fig.~\ref{fig:delta} shows that incorrect answers contain a higher fraction of positive-IS tokens than correct answers, and this gap generally widens with the disagreement threshold. This observation motivates concentrating negative supervision on positions with stronger teacher disagreement. However, increasing the threshold also reduces candidate coverage. In the disagreement-only comparison, $\delta=0.5$ balances corrective value and coverage and achieves the best average accuracy (Table~\ref{tab:delta_margin}(a)).

In the disagreement-only ($\delta$-only) ablation, all tokens receive positive supervision, and every token selected by $g^{(n)}$ is pushed away from the negative teacher. With the sampling distribution of Eq.~\ref{eq:negative}, the objective is
\begin{equation}
\resizebox{0.88\linewidth}{!}{$\displaystyle
\mathcal L_{\delta\text{-only}}(\theta)=\mathcal L_{\pos}(\theta)
-\lambda_{\negat}\,\mathbb E\!\left[
\frac{\sum\limits_{n=1}^{|y|}g^{(n)}D(p_n^s\,\|\,p_n^-)}
{\sum\limits_{n=1}^{|y|}g^{(n)}+\varepsilon}\right]
$}
\label{eq:delta_only}
\end{equation}

\lead{Stage 2: Locating tokens with insufficient positive advantage}
Among the candidate positions, negative supervision should correct predictions closer to the negative teacher and strengthen predictions closer to the positive teacher but without sufficient positive advantage. We define the student--negative-teacher difference and relative difference as
\begin{equation}
\begin{aligned}
d_{s,t^-}^{(n)}&=D\!\left(p_n^s\,\|\,p_n^-\right)\\
d_{s,T}^{(n)}&=d_{s,t^+}^{(n)}-d_{s,t^-}^{(n)}
\end{aligned}
\label{eq:relative}
\end{equation}

The relative difference equals the interference score in Eq.~\ref{eq:is}. A negative value indicates a preference for the positive teacher. A relative preference for the positive teacher does not necessarily imply sufficient separation from the negative teacher. We use the following loss function, with the same sampling distribution as in Eq.~\ref{eq:negative}, to examine how much positive advantage student tokens need:

\begin{equation}
\resizebox{0.88\linewidth}{!}{$\displaystyle
\mathcal L_{m,\,\delta=0.5}(\theta)
= \mathcal L_{\pos}(\theta)
+ \lambda_{\negat}\,\mathbb E\!\left[
\frac{
  \sum\limits_{n=1}^{|y|} g^{(n)}
  \max\!\left(0,\,m+d_{s,T}^{(n)}\right)
}{
  \sum\limits_{n=1}^{|y|} g^{(n)}+\varepsilon
}
\right]
$}
\label{eq:margin_fixed_delta}
\end{equation}

\colorlet{PNsectiongray}{PNgroup}
\definecolor{PNrowpurple}{HTML}{E8E5FF}
\begin{table}[t]
\caption{Accuracy (\%) under different teacher-disagreement
thresholds and margins}
\label{tab:delta_margin}
\centering
\PNtableonefont
\setlength{\tabcolsep}{2pt}
\renewcommand{\arraystretch}{1.18}

\begin{tabular}{@{}
>{\centering\arraybackslash}m{\dimexpr0.28\columnwidth-2\tabcolsep\relax}
>{\centering\arraybackslash}m{\dimexpr0.14\columnwidth-2\tabcolsep\relax}
>{\centering\arraybackslash}m{\dimexpr0.19\columnwidth-2\tabcolsep\relax}
>{\centering\arraybackslash}m{\dimexpr0.25\columnwidth-2\tabcolsep\relax}
>{\centering\arraybackslash}m{\dimexpr0.14\columnwidth-2\tabcolsep\relax}
@{}}
\toprule
Configuration
& V* Bench
& ZoomBench
& \shortstack{MME-RealWorld\\-Lite}
& Avg \\
\midrule

\rowcolor{PNsectiongray}
\multicolumn{5}{c}{
(a) Teacher-disagreement selection only
} \\

All tokens
& 62.30
& \underline{43.79}
& 46.27
& \underline{50.79} \\

$\delta=0$
& \underline{65.45}
& \textbf{43.91}
& 39.71
& 49.69 \\

$\delta=0.2$
& \textbf{69.11}
& 38.11
& 37.47
& 48.23 \\

\rowcolor{PNrowpurple}
$\delta=0.5$
& 62.83
& 41.18
& \textbf{50.03}
& \textbf{51.35} \\

$\delta=0.6$
& 56.02
& 38.93
& \underline{49.50}
& 48.15 \\

\midrule
\rowcolor{PNsectiongray}
\multicolumn{5}{c}{
(b) Margin ablation with $\delta=0.5$
} \\

Disagreement only
& 62.83
& 41.18
& 50.03
& 51.35 \\

$m=0.2$
& 68.59
& 43.43
& 47.94
& 53.32 \\

$m=0.3$
& 67.54
& \underline{44.26}
& \underline{51.43}
& \underline{54.41} \\

$m=0.4$
& 67.54
& 43.43
& 50.55
& 53.84 \\

\rowcolor{PNrowpurple}
$m=0.5$
& \textbf{71.73}
& \textbf{45.44}
& \textbf{52.16}
& \textbf{56.44} \\

$m=0.6$
& \underline{70.68}
& 40.83
& 51.22
& 54.24 \\

$m=0.7$
& 40.31
& 35.50
& 40.07
& 38.63 \\

\bottomrule
\end{tabular}
\end{table}

\noindent\textbf{How Much Positive Advantage Do Student Tokens Need?}\space
\label{sec:margin}
An intuitive choice is to constrain only tokens favoring the negative teacher, whereas our experiments support extending supervision to some tokens already favoring the positive teacher but lacking sufficient positive advantage. As shown in Table~\ref{tab:delta_margin}(b), moderately increasing the required positive advantage helps improve performance, with $m=0.5$ achieving the highest accuracy. An appropriate margin extends negative supervision beyond tokens favoring the negative teacher to include tokens that favor the positive teacher but lack sufficient positive advantage. An excessively large margin, however, may include tokens already exhibiting sufficient positive advantage, imposing unnecessary negative constraints and consequently impairing prediction performance.

\subsection{Optimization Objective}
The two-stage selection is incorporated into a negative distillation objective:
\begin{equation}
\resizebox{0.88\linewidth}{!}{$\displaystyle
\mathcal{L}_{\mathrm{negative}}(\theta)
=
\mathbb{E}_{\substack{
(I,q,R)\sim\mathcal{D}\\
y\sim p^s(\cdot\mid I,q)
}}
\left[
\frac{
\sum\limits_{n=1}^{|y|}g^{(n)}
\max\!\left(0,m+d_{s,T}^{(n)}\right)
}{
\sum\limits_{n=1}^{|y|}g^{(n)}+\varepsilon
}
\right]
$}
\label{eq:negative}
\end{equation}
where $m\geq0$ is the positive-advantage margin and $\varepsilon>0$ prevents division by zero. The hinge term activates only when $m+d_{s,T}^{(n)}>0$. A token has sufficient positive advantage once
\begin{equation}
d_{s,t^-}^{(n)}\geq d_{s,t^+}^{(n)}+m
\label{eq:advantage}
\end{equation}
In other words, the student's difference to the negative teacher should exceed its difference to the positive teacher by at least $m$. Otherwise, negative distillation remains active at candidate positions. The repulsion is therefore anchored to the positive teacher, rather than maximizing the absolute difference to the negative teacher without a reference.

The overall training objective is
\begin{equation}
\resizebox{0.88\linewidth}{!}{$\displaystyle
\mathcal L(\theta)=\mathcal L_{\pos}(\theta)
+\lambda_{\negat}\mathcal L_{\negat}(\theta)
\quad \lambda_{\negat}>0
$}
\label{eq:total}
\end{equation}
Positive distillation continuously aligns all valid positions with evidence-supported predictions, while negative distillation constrains selected positions affected by the negative teacher or lacking sufficient positive advantage. The negative loss is zero at positions failing the activation conditions.

\section{Experiments}
\label{sec:experiments}

\subsection{Experimental Setup}
\subsubsection{Benchmarks}
Evaluation covers three benchmarks for Fine-Grained Visual Tasks, with POPE additionally used to assess object hallucination. V* Bench~\cite{r06} measures the accuracy of locating and recognizing small visual targets in complex scenes. ZoomBench~\cite{r09} contains questions whose answers depend on visual details at different zoom levels. MME-RealWorld-Lite~\cite{r18} covers real-world scenarios with high-resolution images. POPE~\cite{r25} evaluates the extent of object hallucination in multimodal large language models.

\subsubsection{Model training}
SPN-OPSD is trained on the 5.8K Vision-OPD~\cite{r10} examples with fewer than 4,096 input tokens, using Qwen3.5-0.8B and Qwen3.5-2B.

\subsubsection{Baselines}
We conduct two groups of comparisons.  The first compares SPN-OPSD with existing multimodal models: (a) closed-source MLLMs, including GPT-5.2~\cite{r19} and GPT-5.4~\cite{r20}; (b) open-source MLLMs, including Qwen3.5~\cite{r21}, MiniCPM-V-4.5~\cite{r22}, and MiMo-V2.5~\cite{r23}; and (c) thinking-with-images agents, including DeepEyes V2~\cite{r24} and Thyme~\cite{r08}.

The second group compares training strategies for fine-grained visual tasks under the same training data and backbone. DeFacto~\cite{r11} uses evidence localization, CompoDistill~\cite{r13} uses attention alignment, and VA-OPD~\cite{r15} adjusts distillation weights with counterfactual signals. Table~\ref{tab:main} reports the comparison results.

\definecolor{PNgroup}{HTML}{F2F2F2}
\definecolor{PNpurple}{HTML}{E8E5FF}
\definecolor{PNgain}{HTML}{B84DE3}
\begin{table}[t]
\caption{Comparison with multimodal models}
\label{tab:main}
\centering
\scriptsize
\setlength{\tabcolsep}{3pt}
\renewcommand{\arraystretch}{1.08}

\sbox{\PNmaintablebox}{%
\begin{tabular}{@{}lccccc@{}}
\toprule
\multicolumn{1}{@{}c}{Model}
& Size
& V* Bench
& ZoomBench
& \makecell[c]{MME-\\RealWorld-Lite}
& Avg \\
\midrule

\rowcolor{PNgroup}
\multicolumn{6}{c}{\textit{Visual agents}} \\
DeepEyes V2 & 7B & 81.68 & 44.97 & 51.59 & 59.41 \\
Thyme & 7B & 82.20 & 45.09 & 55.18 & 60.82 \\
\midrule

\rowcolor{PNgroup}
\multicolumn{6}{c}{\textit{Closed-source models}} \\
GPT-5.2 & -- & 79.06 & 50.89 & 47.06 & 59.00 \\
GPT-5.4 & -- & 76.96 & \underline{52.66} & 48.10 & 59.24 \\
\midrule

\rowcolor{PNgroup}
\multicolumn{6}{c}{\textit{Open-source models}} \\
Qwen3.5 & 4B & \underline{84.29} & 47.69 & 35.80 & 55.93 \\
Qwen3.5 & 9B & 82.72 & 52.07 & 46.12 & 60.30 \\
MiniCPM-V-4.5 & 9B & 70.68 & 42.60 & 42.31 & 51.86 \\
MiMo-V2.5 & 310B & 82.20 & 51.24 & 46.69 & 60.04 \\
\midrule

\rowcolor{PNgroup}
\multicolumn{6}{c}{\textit{Matched backbone and training data}} \\
Qwen3.5 & 0.8B & 66.49 & 40.71 & 50.65 & 52.62 \\
+ VA-OPD & 0.8B & 64.40 & 39.29 & 46.95 & 50.21 \\
+ DeFacto & 0.8B & 67.54 & 42.84 & 51.75 & 54.04 \\
+ CompoDistill & 0.8B & 68.59 & 41.66 & 52.79 & 54.35 \\

\rowcolor{PNpurple}
+ SPN-OPSD (ours) & 0.8B & 71.73 & 45.44 & 52.16 & 56.44 \\
\rowcolor{PNpurple}
& & \textcolor{PNgain}{↑5.24}
  & \textcolor{PNgain}{↑4.73}
  & \textcolor{PNgain}{↑1.51}
  & \textcolor{PNgain}{↑3.82} \\

Qwen3.5 & 2B & 76.44 & 45.92 & 57.43 & 59.93 \\
+ VA-OPD & 2B & 83.25 & 43.55 & 53.26 & 60.02 \\
+ DeFacto & 2B & 79.58 & 49.23 & \underline{58.99} & 62.60 \\
+ CompoDistill & 2B & 82.72 & 50.53 & 58.10 & \underline{63.78} \\

\rowcolor{PNpurple}
+ SPN-OPSD (ours) & 2B
& \textbf{84.82}
& \textbf{54.56}
& \textbf{59.56}
& \textbf{66.31} \\
\rowcolor{PNpurple}
& & \textcolor{PNgain}{↑8.38}
  & \textcolor{PNgain}{↑8.64}
  & \textcolor{PNgain}{↑2.13}
  & \textcolor{PNgain}{↑6.38} \\
\bottomrule
\end{tabular}%
}
\PNrecordtablefont
\resizebox{\columnwidth}{!}{\usebox{\PNmaintablebox}}
\end{table}

\flushbottom
\subsection{Main Results}
As shown in Table~\ref{tab:main}, SPN-OPSD consistently outperforms the corresponding Qwen3.5-0.8B and Qwen3.5-2B base models across all three benchmarks, indicating successful internalization of the ability to reason from local evidence. At 2B it also exceeds the larger open-source models, GPT-5.4, and the thinking-with-images agents while using standard full-image decoding without inference-time tools. At both model sizes, SPN-OPSD achieves the highest average accuracy among the compared training strategies for fine-grained visual tasks. The results support the effectiveness of combining positive distillation with selective negative supervision.

\section{Experimental Analysis}
\label{sec:analysis}

The present section answers the following question: Does Selective Negative Teacher Supervision Provide Additional Benefits?

\label{sec:components}
\subsection{Both Components Are Necessary}

To determine whether the negative teacher provides gains beyond positive distillation and whether the gains depend on supervision positions, we compare six configurations at both model sizes. Table~\ref{tab:ablation} shows full SPN-OPSD achieves the best overall performance. Applying negative constraints to all tokens degrades performance on some benchmarks. The negative teacher therefore provides additional information beyond positive supervision, but selective use is necessary for consistent gains. Removing either selection stage reduces performance relative to full SPN-OPSD, indicating partial recovery of the improvement on some benchmarks with single-stage configurations, whereas combining both stages yields the full benefit.

\begin{table}[t]

\providecolor{PNgroup}{HTML}{F2F2F2}
\colorlet{ABgroup}{PNgroup}
\definecolor{ABpurple}{HTML}{E8E4FF}
\definecolor{ABgain}{HTML}{B84DE3}

\caption{Component ablation on Qwen3.5-0.8B and Qwen3.5-2B}
\label{tab:ablation}
\centering
\PNtableonefont
\setlength{\tabcolsep}{2.5pt}
\renewcommand{\arraystretch}{1.15}

\def\abscore#1#2{%
  \begin{tabular}[c]{@{}c@{}}
    \textbf{#1}\\[-0.45ex]
    {\scriptsize\textcolor{ABgain}{$\uparrow\,#2$}}%
  \end{tabular}%
}

\begin{tabular}{@{}
  w{c}{\dimexpr0.32\columnwidth-\tabcolsep\relax}
  w{c}{\dimexpr0.12\columnwidth-2\tabcolsep\relax}
  w{c}{\dimexpr0.165\columnwidth-2\tabcolsep\relax}
  w{c}{\dimexpr0.195\columnwidth-2\tabcolsep\relax}
  w{c}{\dimexpr0.10\columnwidth-2\tabcolsep\relax}
  w{c}{\dimexpr0.10\columnwidth-\tabcolsep\relax}
@{}}
\toprule
Configuration
& \begin{tabular}[c]{@{}c@{}}V*\\Bench\end{tabular}
& ZoomBench
& \begin{tabular}[c]{@{}c@{}}MME-\\RealWorld-Lite\end{tabular}
& POPE
& Avg \\
\midrule

\rowcolor{ABgroup}
\multicolumn{6}{c}{\textbf{Qwen3.5-0.8B}} \\

Base
& 66.49 & 40.71 & 50.65 & 86.50 & 61.09 \\

\begin{tabular}[c]{@{}c@{}}Positive-only\\distillation\end{tabular}
& 67.02 & 40.83 & 50.91 & 84.31 & 60.77 \\

All-token SPN-OPSD
& 62.30 & 43.79 & 46.27 & 84.87 & 59.31 \\

\begin{tabular}[c]{@{}c@{}}Teacher-disagreement\\selection only\end{tabular}
& 62.83 & 41.18 & 50.03 & 86.79 & 60.21 \\

\begin{tabular}[c]{@{}c@{}}Relative-difference\\selection only\end{tabular}
& 69.63 & 41.18 & 48.62 & 86.52 & 61.49 \\

\rowcolor{ABpurple}
\textbf{SPN-OPSD (ours)}
& \abscore{71.73}{2.10}
& \abscore{45.44}{1.65}
& \abscore{52.16}{1.25}
& \abscore{86.91}{0.12}
& \abscore{64.06}{2.57} \\

\midrule
\rowcolor{ABgroup}
\multicolumn{6}{c}{\textbf{Qwen3.5-2B}} \\

Base
& 76.44 & 45.92 & 57.43 & 89.31 & 67.28 \\

\begin{tabular}[c]{@{}c@{}}Positive-only\\distillation\end{tabular}
& 76.96 & 50.89 & 49.87 & 88.73 & 66.61 \\

All-token SPN-OPSD
& 80.10 & 51.48 & 56.59 & 88.16 & 69.08 \\

\begin{tabular}[c]{@{}c@{}}Teacher-disagreement\\selection only\end{tabular}
& 83.25 & 51.36 & 57.06 & 89.42 & 70.27 \\

\begin{tabular}[c]{@{}c@{}}Relative-difference\\selection only\end{tabular}
& 82.72 & 52.78 & 57.22 & 88.47 & 70.30 \\

\rowcolor{ABpurple}
\textbf{SPN-OPSD (ours)}
& \abscore{84.82}{1.57}
& \abscore{54.56}{1.78}
& \abscore{59.56}{1.93}
& \abscore{89.84}{0.42}
& \abscore{72.20}{1.90} \\

\bottomrule
\end{tabular}
\end{table}

\subsubsection{Gains from Teacher Disagreement Filtering Stem from Effective Selection of Supervision Positions}
Because changing $\delta$ alters both the number and locations of selected tokens, we add a count-matched random-selection control. Teacher-disagreement selection outperforms random selection on all three benchmarks (Table~\ref{tab:selection_controls}(a)), confirming that disagreement identifies more informative positions rather than merely selecting tokens.
\subsubsection{Gains from token selection based on student relative difference}
Varying $m$ changes both supervision count and positions, so we again use a count-matched random control; relative-difference selection beats random selection on all three benchmarks (Table~\ref{tab:selection_controls}(b)).

\subsubsection{Contributions of negative constraints}

The active negative hinge $\max(0,m+d_{s,t^+}^{(n)}-d_{s,t^-}^{(n)})$ encourages both separation from the negative teacher and alignment with the positive teacher. The observed gains could therefore partly arise from stronger positive supervision at the selected positions. To test the additional value of the repulsive constraint, we introduce a positive-upweighting control. The control retains teacher-disagreement selection and margin-based activation but replaces the active negative term with the student--positive-teacher divergence. Specifically,
\begin{equation}
\mathcal{L}_{\mathrm{auxiliary}}(\theta)
=
\mathbb{E}_{\substack{
(I,q,R)\sim\mathcal{D}\\
y\sim p^s(\cdot\mid I,q)
}}
\left[
\frac{
\sum\limits_{n=1}^{|y|}h^{(n)}d_{s,t^+}^{(n)}
}{
\sum\limits_{n=1}^{|y|}g^{(n)}+\varepsilon
}
\right]
\label{eq:aux}
\end{equation}
\begin{equation}
h^{(n)}
=
g^{(n)}\ind\!\left[
m+d_{s,t^+}^{(n)}-d_{s,t^-}^{(n)}>0
\right]
\label{eq:active_mask}
\end{equation}
The control optimizes $\mathcal L_{\pos}+\lambda_{\negat}\mathcal L_{\mathrm{aux}}$, with the other settings unchanged. Full SPN-OPSD outperforms positive upweighting on all three benchmarks in Table~\ref{tab:selection_controls}(b). Encouraging the student to move away from the negative teacher therefore provides additional gains beyond increasing positive supervision at the same selected positions.

\colorlet{PNsectiongray}{PNgroup}
\definecolor{PNrowpurple}{HTML}{E8E5FF}
\definecolor{PNgainpurple}{HTML}{B84DE3}

\newcommand{\PNup}[1]{%
  {\scriptsize\textcolor{PNgainpurple}{$\uparrow\,#1$}}%
}
\begin{table}[t]
\caption{Comparison with count-matched random selection and
positive upweighting}
\label{tab:selection_controls}
\centering
\PNtableonefont
\setlength{\tabcolsep}{2pt}
\renewcommand{\arraystretch}{1.18}

\begin{tabular}{@{}
>{\centering\arraybackslash}m{\dimexpr0.38\columnwidth-2\tabcolsep\relax}
>{\centering\arraybackslash}m{\dimexpr0.14\columnwidth-2\tabcolsep\relax}
>{\centering\arraybackslash}m{\dimexpr0.18\columnwidth-2\tabcolsep\relax}
>{\centering\arraybackslash}m{\dimexpr0.18\columnwidth-2\tabcolsep\relax}
>{\centering\arraybackslash}m{\dimexpr0.12\columnwidth-2\tabcolsep\relax}
@{}}
\toprule
Configuration
& \shortstack{V*\\Bench}
& ZoomBench
& \shortstack{MME-\\RealWorld-Lite}
& Avg \\
\midrule

\rowcolor{PNsectiongray}
\multicolumn{5}{c}{(a) Teacher-disagreement selection} \\

\shortstack{Count-matched\\random}
& 82.20 & 50.77 & 55.08 & 62.68 \\

\rowcolor{PNrowpurple}
& \textbf{83.25}
& \textbf{51.36}
& \textbf{57.06}
& \textbf{63.89} \\

\rowcolor{PNrowpurple}
\multirow{-2}{*}{Disagreement ($\delta=0.5$)}
& \PNup{1.05}
& \PNup{0.59}
& \PNup{1.98}
& \PNup{1.21} \\

\midrule
\rowcolor{PNsectiongray}
\multicolumn{5}{c}{(b) Full two-stage selection} \\

\shortstack{Count-matched\\random}
& 84.29 & 50.89 & 56.59 & 63.92 \\

\shortstack{Positive\\upweighting}
& 80.10 & 48.88 & 57.43 & 62.14 \\

\rowcolor{PNrowpurple}
& \textbf{84.82}
& \textbf{54.56}
& \textbf{59.56}
& \textbf{66.31} \\

\rowcolor{PNrowpurple}
\multirow{-2}{*}{SPN-OPSD(ours)}
& \PNup{0.53}
& \PNup{3.67}
& \PNup{2.13}
& \PNup{2.39} \\

\bottomrule
\end{tabular}
\end{table}


\subsection{SPN-OPSD Improves the Use of Local Visual Evidence}

Models often err on Fine-Grained Visual Tasks because they struggle to locate relevant evidence~\cite{r10}. We assess whether SPN-OPSD alleviates the difficulty in the following three ways.

\begin{figure}[t]
\centering
\includegraphics[width=\columnwidth]{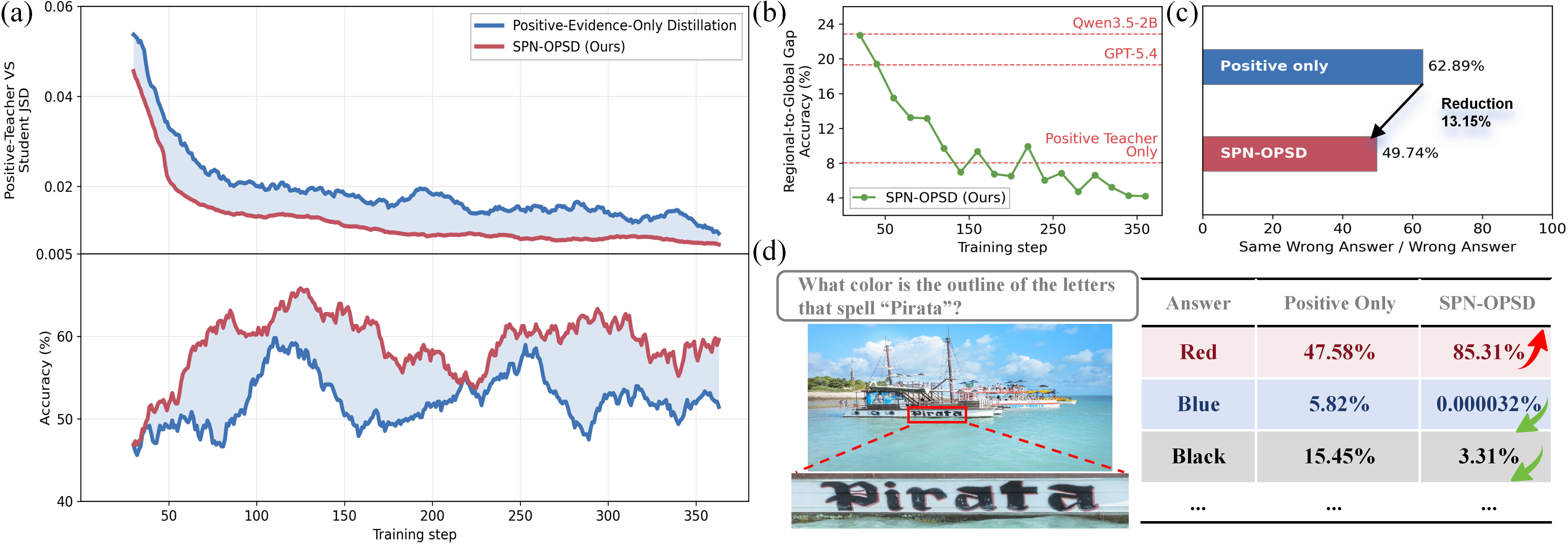}
\caption{Comparison between positive-only distillation and SPN-OPSD. In Fig. 4(a), the curves are smoothed using a 30-step moving average and start at step 30.}
\label{fig:training}
\end{figure}

In Fig.~\ref{fig:training}(a), SPN-OPSD yields lower student--positive-teacher JSD (top) and higher answer accuracy (bottom) than positive-only distillation over most of training. Under selective negative supervision, closer alignment thus accompanies higher accuracy. Fig.~\ref{fig:training}(d) illustrates how the gain is reflected in the student's predictions. Asked about the outline color of ``Pirata,'' the SPN-OPSD student assigns greater probability to the correct answer, ``red,'' than its positive-only counterpart. It assigns almost none to ``blue,'' the base model's answer after the evidence region is blurred. SPN-OPSD improves accuracy by increasing confidence in correct answers and assigning less probability to incorrect alternatives.

In Fig.~\ref{fig:training}(b), the accuracy gap between full-image and cropped-image inputs generally narrows during training. The final gap is smaller for SPN-OPSD than for positive-only distillation, the base model, or GPT-5.4, suggesting more effective use of local evidence in the full image.

Fig.~\ref{fig:training}(c) compares predictions before and after blurring only the evidence region. For each model, we consider questions answered incorrectly in both views. SPN-OPSD repeats the same wrong answer less often than positive-only distillation, suggesting less reliance on unchanged background cues and a greater ability to resist interference from irrelevant regions.

\subsection{Performance with very small evidence regions}
To examine model sensitivity to evidence-region size, we group the 845 ZoomBench examples by the ratio of evidence-box area to image area. Table~\ref{tab:size} shows SPN-OPSD outperforms both the base model and positive-only distillation across all three groups. The largest gain over the base model occurs in the smallest-region group, supporting the advantage of SPN-OPSD on tasks requiring minute visual details.

\begin{table}[t]
\caption{Accuracy (\%) on ZoomBench for different evidence-region sizes}
\label{tab:size}
\centering
\scriptsize
\setlength{\tabcolsep}{3pt}
\renewcommand{\arraystretch}{1.25}

\begin{tabular}{
    >{\centering\arraybackslash}m{0.8cm}
    >{\centering\arraybackslash}m{1.3cm}
    >{\centering\arraybackslash}m{0.9cm}
    >{\centering\arraybackslash}m{1.2cm}
    >{\centering\arraybackslash}m{1.5cm}
    >{\raggedright\arraybackslash}m{1.6cm}
}
\toprule
\shortstack{Evidence\\Region}
& \shortstack{Area\\Ratio}
& \shortstack{Sample\\Count}
& \shortstack{Qwen3.5-\\2B}
& \shortstack{Positive-Only\\Distillation}
& {\centering SPN-OPSD\par} \\
\midrule
Small
& $<2.5\%$
& 286
& 39.51
& 55.94
& \cellcolor{PNlight}\mbox{\textbf{58.04}\,\gain{18.53}} \\

Medium
& $[2.5,5.5)\%$
& 278
& 49.64
& 52.16
& \cellcolor{PNlight}\mbox{\textbf{53.96}\,\gain{4.32}} \\

Large
& $\geq5.5\%$
& 281
& 48.75
& 44.48
& \cellcolor{PNlight}\mbox{\textbf{51.60}\,\gain{2.85}} \\

Overall
& --
& 845
& 45.92
& 50.89
& \cellcolor{PNlight}\mbox{\textbf{54.56}\,\gain{8.64}} \\
\bottomrule
\end{tabular}
\end{table}

\section{Conclusion}
This paper proposes SPN-OPSD, which uses both evidence and background regions to provide supervision and improve the model's resistance to interference from irrelevant regions. By combining continuous positive distillation with selective negative distillation, SPN-OPSD learns evidence-supported distributions while identifying token-level positions with high value for negative supervision, enabling targeted distillation. We hope SPN-OPSD can advance the recognition and reasoning capabilities of multimodal large language models in fine-grained visual tasks.

\end{document}